\documentclass[11pt,a4paper]{article}
\usepackage[hyperref]{acl2020}
\usepackage{times}
\usepackage{latexsym}

\usepackage{microtype}
\usepackage{multirow}
\usepackage{caption}
\usepackage{subcaption}
\usepackage{graphicx} 
\usepackage{booktabs}
\usepackage{trackchanges}

\addeditor{YJ}

\newcommand\shortsectionnp[1]{\vspace{6pt}{\noindent\bf #1}}

\aclfinalcopy 

\title{Pointwise Paraphrase Appraisal is Potentially Problematic}

\author{Hannah Chen, 
    Yangfeng Ji, David Evans \\
    Department of Computer Science\\
    University of Virginia\\
    Charlottesville, VA 22904\\
  \texttt{\{yc4dx,yangfeng,evans\}@virginia.edu} \\}

\date{}

\begin{document}
\maketitle
\begin{abstract}
The prevailing approach for training and evaluating paraphrase identification models is constructed as a binary classification problem: the model is given a pair of sentences, and is judged by how accurately it classifies pairs as either paraphrases or non-paraphrases. This pointwise-based evaluation method does not match well the objective of most real world applications, so the goal of our work is to understand how models which perform well under pointwise evaluation may fail in practice and find better methods for evaluating paraphrase identification models. As a first step towards that goal, we show that although the standard way of fine-tuning BERT for paraphrase identification by pairing two sentences as one sequence results in a model with state-of-the-art performance, that model may perform poorly on simple tasks like identifying pairs with two identical sentences. Moreover, we show that these models may even predict a pair of randomly-selected sentences with higher paraphrase score than a pair of identical ones.
\end{abstract}

\section{Introduction}

Paraphrase identification is a well-studied sentence pair modeling task that refers to the problem of determining whether two sentences are semantically equivalent. Detecting paraphrases can be very useful for many NLP applications such as machine translation (MT), question answering (QA), and information retrieval (IR). In a QA system, we would like to find the most probable question paraphrases from a database of question answer pairs for a given input question~\citep{paraphrase-in-qa, dong-etal-2017-learning}. In a MT model, we would like to obtain the best translation by comparing the target sentence to a list of translated sentences. Even though pre-trained language models have reached state-of-the-art performance on paraphrase identification tasks, the current problem setup is insufficient to produce models with consistent and robust performance on unseen samples and real world problems.

The typical current problem setup for paraphrase identification is different from intended uses in real world applications. They often involve finding best paraphrases from a group of documents given a particular query, rather than just determining whether two sentences are paraphrases of each other. Besides, getting the order and identifying the most relevant documents is usually more important than getting the binary decision of a pair of sentences~\citep{top-k-retrieval}. However, to make the task simpler, current methods and existing datasets such as Quora Question Pairs (QQP) \citep{qqp} and Microsoft Research Paraphrase Corpus (MRPC) \citep{lan-qiu-2017-continously} are all framed as a binary classification problem at the sentence pair level. 

\shortsectionnp{Contributions}\hspace{0.25cm}
As a first step to improve the way paraphrase identification is evaluated for ranking tasks, we analyze some of the anomalies found in the current pointwise task setting. We first demonstrate the standard way of fine-tuning BERT for pointwise paraphrase evaluation makes the model sometimes fails on simple problems including recognizing two identical sentences and reversing the order of two sentences in a pair (Section~\ref{asymmetry_in_bert}). We find that it performs worse than a bag-of-words model due to its asymmetrical model architecture. Lastly, we show that the model may fail to capture the correct relative order of two sentence pairs using the pointwise approach, sometimes even predicting a pair of random sentences with a higher paraphrase score than a pair of identical ones (Section~\ref{pointwise_problems}).

\section{Background}

This section provides background on the paraphrase identification task, evaluation methods, and the datasets and models we use in our experiments.

\subsection{Paraphrase Identification}
We consider the general definition of paraphrase as sentences having the same meaning. In addition, paraphrase requires a symmetric relation. Paraphrase identification originates from the real-world applications such as machine translation \citep{unsupervised-paraphrase-corpora,quirk2004monolingual} and document summarization \citep{barzilay2001extracting}, where an essential task is to evaluate the semantic relatedness of translated sentences or generated texts. 

\subsection{Evaluation Methods}

The current problem setting for paraphrase identification is similar to the pointwise method for learning-to-rank problems in information retrieval~\citep{l2r}. There are three types of approaches to solve learning-to-rank: pointwise, pairwise, and listwise~\citep{l2r-for-IR}. The \textit{pointwise} approach learns to predict a binary relevance judgement for a single document given a specific query. It retrieves the most relevant document by computing the relevance score between each candidate document and the query and returning the document with the maximum score. The \textit{pairwise} approach learns to predict the relative order of a pair of documents, $(d_1,d_2)$, for a given query $q$. This is closer to the nature of ranking than the pointwise approach. However, both the pointwise and pairwise approaches neglect the fact that some documents are related to the same query. The \textit{listwise} approach directly optimizes the model on the permutations of a list of documents $D=\{d_1, d_2,...,d_n\}$~\citep{l2r-listwise}, and hence it most closely matches the objective of ranking.

\subsection{Datasets}
\label{datasets}

For our experiments, we use four datasets designed for evaluating paraphrase identification models.

\shortsectionnp{Quora Question Pairs (QQP)} consists of 400k question pairs from Quora~\citep{qqp}. The goal is to reduce the number of duplicate questions on the platform. Each question pair is either labeled as duplicate or non-duplicate. Recently, it has been shown to have selection bias, where models can simply rely on the frequency of the sentences or the intersection of the neighbor sentences to make predictions \citep{selection_bias}.

\shortsectionnp{Paraphrase Adversaries from Word Scrambling (PAWS)} contains two datasets constructed from Wikipedia and QQP~\citep{PAWS}. To compare with the original QQP dataset, we only tested PAWS\textsubscript{QQP}. The sentence pairs are created by swapping words that have the same part-of-speech or named entity tags to construct higher lexical overlap sentences. The training set contains 11,988 sentence pairs, and the testing set contains 667.

\shortsectionnp{Microsoft Research Paraphrase Corpus (MRPC)} contains 5801 sentence pairs extracted from online news articles~\citep{dolan2005automatically}. The sentence pairs are created with very similar syntactic features and high \emph{n}-gram overlap causing the model to make skewed decisions based on these shallow features~\citep{das-smith-paraphrase}.  

\shortsectionnp{Twitter URL Paraphrase Corpus} is extracted from tweets posted by 22 English news accounts on Twitter~\citep{lan-qiu-2017-continously}. Relevant tweets are paired up based on the same embedded URLs, and each pair is then labeled by 6 human annotators. After discarding sentence pairs with neutral decisions (3 out of 6 annotators labeled it as paraphrase), the dataset consists of 42k sentence pairs for training and 9k pairs for testing. 

\begin{table}[bt]
\centering
\resizebox{\columnwidth}{!}{%
    \begin{tabular}{l c c c c}
    \toprule
        & \multicolumn{2}{c}{BERT} & \multicolumn{2}{c}{BOW} \\
    \multicolumn{1}{c}{Dataset} & Acc & F1 & Acc & F1 \\
    \midrule
    QQP & 90.10 & 86.7 & 64.75 & 51.56 \\
    QQP+PAWS\textsubscript{QQP} & 90.69 & 87.48 & 64.13 & 51.28 \\
    MRPC & 83.65 & 87.97 & 68.12 & 79.45 \\
    Twitter URL & 89.98 & 76.75 & 84.32 & 50.44 \\
    \bottomrule
    \end{tabular}%
}
\caption{Model accuracy and F1 scores trained on different datasets. Both metrics are scaled by 100. QQP + PAWS\textsubscript{QQP} indicates models are trained and evaluated on both datasets.}
\label{model_acc_f1}
\end{table}

\subsection{Models}
We fine-tuned the BERT\textsubscript{BASE} model on different paraphrase datasets with the default configuration~\citep{BERT}. We also implemented early stopping during the training process. For baseline comparison, we trained a bag-of-words (BOW) model with unigram and bigram encodings. The model makes predictions based on the consine similarity between the encodings of the two sentences. A consine similarity value above 0.5 is considered a paraphrase. The performance of both models for each task is shown in Table~\ref{model_acc_f1}. 

We include the results for testing QQP model on its adversarial set, PAWS\textsubscript{QQP}, in Table~\ref{adversarial}, and it shows BERT performing as poorly as BOW of this dataset. We also report the results of models that trained and tested on a concatenated set of QQP and PAWS\textsubscript{QQP} in Table~\ref{model_acc_f1}.

\begin{table}[tb]
\centering
\begin{tabular}{l c c}
    \toprule
    \multirow{2}{4em}{Models} & \multicolumn{2}{c}{QQP $\rightarrow$ PAWS\textsubscript{QQP}}\\
     & Acc & F1 \\
    \midrule
    BERT & 32.94 & 42.68 \\
    BOW & 28.21 & 44.01 \\
    \bottomrule
\end{tabular}
\caption{Model accuracy and F1 score tested in the adversarial setting, where models are trained on QQP and evaluated on PAWS\textsubscript{QQP} development set.}
\label{adversarial}
\end{table}

\section{Asymmetry}
\label{asymmetry_in_bert}
For semantic matching tasks, the BERT paraphrase identification model considers two sentences $(s_1, s_2)$ as a single sequence by concatenating them with a separator token. However, due to this asymmetrical approach, the sequence representations before the final classification layer would be entirely different if we permute the order of the two input sentences. We explore two implications of this method for identifying paraphrases: sensitivity to input order (Section~\ref{sentence_order}) and possibility of considering identical sentences non-paraphrases (Section~\ref{identifying_identical_pairs}).


\subsection{Sensitivity to Sentence Order}
\label{sentence_order}
In the original datasets, each sentence pair is only concatenated in one way as $(s_1, s_2)$ and a label $y$ will be predicted by the model. We constructed new sentence pairs in the reverse order as $(s_2, s_1)$, and tested the model on these sentence pairs and got their predicted labels $y^{\prime}$. To find out how much it would affect the prediction results, we computed the ratio of sentence pairs that are predicted with a different label ($y\neq y^{\prime}$). The results for BERT and BOW models are shown in the second and third column of Table~\ref{asymmetry_ratio}.

In normal setting (model is trained and evaluated on the the same dataset), there are more than 3\% of sentence pairs that are predicted with an opposite label by BERT. The ratio decreases on PAWS\textsubscript{QQP}, but it increases when the model includes adversarial examples in the training data. The percentages are even higher on MRPC and Twitter corpus. BOW, trivially, has zero disagreement since the order does not effect the bag-of-words model.

We reproduced the same experiment in section~\ref{sentence_order} on the RoBERTa\textsubscript{BASE} model \citep{roberta}, and found that the model also has inherent asymmetry issue as BERT. The ratio of sentence pairs from the QQP development set with opposite labels is around 4.7\% (But it performs well on identifying identical sentences with an error rate less than 1\%). We further tried fine-tuning BERT on the augmented QQP dataset that includes sentence pairs in both original and reverse order. Although the ratio of sentence pairs with opposite predicted labels decreases about half, the asymmetrical issue is not completely eliminated. These results suggest that these pre-trained language model do not really understand the symmetric relation within paraphrases. One possible reason is combining two sentences as a single input encourages the model to learn paraphrase as an asymmetric relation.

\begin{table}[htbp!]
\centering
\resizebox{\columnwidth}{!}{%
    \begin{tabular}{lcccc}
    \toprule
    \multirow{2}{4em}{Datasets} & \multicolumn{2}{c}{Reverse Order} & \multicolumn{2}{c}{Identical}\\
     & BERT & BOW & BERT & BOW \\
    \midrule
    QQP$\rightarrow$QQP & 3.70 & 0.0 & 2.40 & 0.0  \\
    QQP$\rightarrow$PAWS\textsubscript{QQP} & 2.66 & 0.0 & 7.36 & 0.0 \\
    QQP+PAWS\textsubscript{QQP} & 4.0 & 0.0 & 0.54 & 0.0 \\
    MRPC & 8.46 & 0.0 & 0.0 & 0.0 \\
    Twitter URL & 7.08 & 0.0 & 0.0 & 0.0 \\
    \bottomrule
    \end{tabular}%
}
\caption{The percentage (\%) of sentence pairs with asymmetrical prediction results. Reverse Order: sentence pairs predicted with different labels when reversing the order of the sentences. Identical: identical pairs that are predicted as non-paraphrases. (Please see Section~\ref{datasets} for actual data sizes.)}
\label{asymmetry_ratio}
\end{table}

\begin{table*}[tb]
    \centering
    \resizebox{\linewidth}{!}{%
    \begin{tabular}{l c c c c c}
    \toprule
        & QQP $\rightarrow$ QQP & QQP $\rightarrow$ PAWS\textsubscript{QQP} & QQP + PAWS\textsubscript{QQP} & MRPC & Twitter URL \\
    \midrule
    Paraphrase $>$ Identical & 30.51 & 41.88 & 21.27 & 4.18 & 1.76 \\
    Avg Score Difference & 5.09 & 3.07 & 2.58 & 0.12 & 0.60 \\
     \midrule
    Non-paraphrase $>$ Identical & 0.97 & 43.21 & 0.41 & 0.0 & 0.03 \\
    Avg Score Difference & 6.28 & 2.53 & 3.01 & 0.0 & 1.52 \\
    \bottomrule
    \end{tabular}%
    }
    \caption{Percentage of paraphrase and non-paraphrase pairs with higher paraphrase score (\%) than a pair of identical sentences given the same query sentence. Avg Score Differences: average score difference between paraphrase/non-paraphrase and identical pairs. (Only pairs with higher scores than the identical ones are included.)}
    \label{paraphrase_score}
\end{table*}

\subsection{Inability to Recognize Identical Sentences}
\label{identifying_identical_pairs}
We would like to know if the asymmetrical structure also affect BERT's ability to identify identical sentences as paraphrases. We collected distinct sentences for each dataset and constructed a new set of sentence pairs by pairing each one with itself. Each pair is labeled as paraphrase. We calculated the ratio of pairs that are predicted as non-paraphrase by the model.
As shown in Table~\ref{asymmetry_ratio}, BERT trained on QQP recognizes 2.4\% of identical pairs as non-paraphrases and the ratio increases about 5\% when tested on PAWS\textsubscript{QQP}. BOW trivially achieves perfect accuracy on pairs of identical sentences, since they have exactly the same bags of words.

The models trained on MRPC and Twitter corpus do recognize all the identical pairs as paraphrases. This may be the fact that many sentences appear in Twitter corpus multiple times pairing with different sentence each time. Thus, the model may better capture the difference between a variety of sentences. As for MRPC, many sentence pairs look quite alike, and hence the model can better identify small differences between sentences even though most sentences only appear once. Since PAWS\textsubscript{QQP} contains higher lexical overlap sentence pairs, the model trained on both QQP and PAWS\textsubscript{QQP} decreases the error rate to less than 1\%. 

We also fine-tuned a BERT model on the augmented QQP training set with identical sentence pairs, and it can correctly identify every identical pairs as paraphrases. This suggests that the amount of lexical overlap in the dataset would affect the model's ability to identify identical sentences. 



\section{Problems with Pointwise Evaluation}
\label{pointwise_problems}
For a given query sentence, we assume that a well-generalized paraphrase identification model should output a higher paraphrase score to the query sentence itself than a randomly-selected sentence. However, models trained with pointwise evaluation cannot learn the relative order based on the degree of semantic equivalence. We test this by considering how often models recognize a random sentence as more similar than the query sentence itself, and looking at the distribution of paraphrase scores across a dataset.

\begin{figure*}[tb]
\centering

\begin{subfigure}[b]{1.0\textwidth}
   \includegraphics[width=1\linewidth]{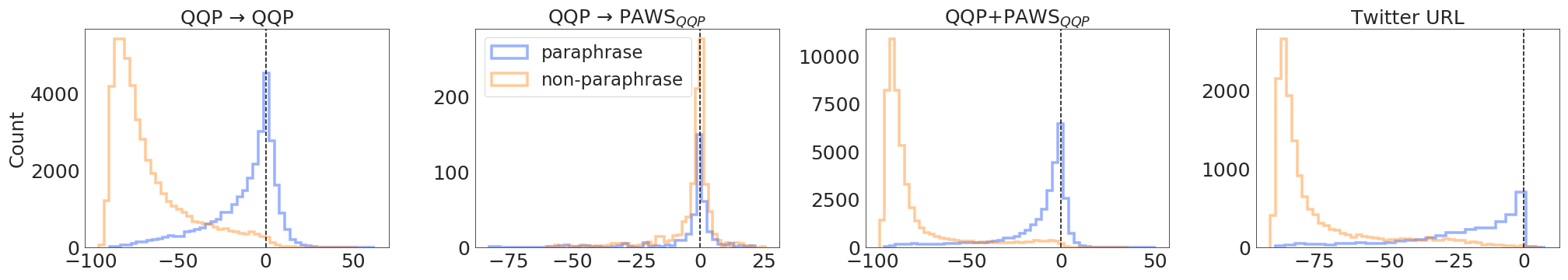}
   \caption{}
   \label{confidence_diff_hist}
\end{subfigure}

\begin{subfigure}[b]{1.0\textwidth}
    \includegraphics[width=1\linewidth]{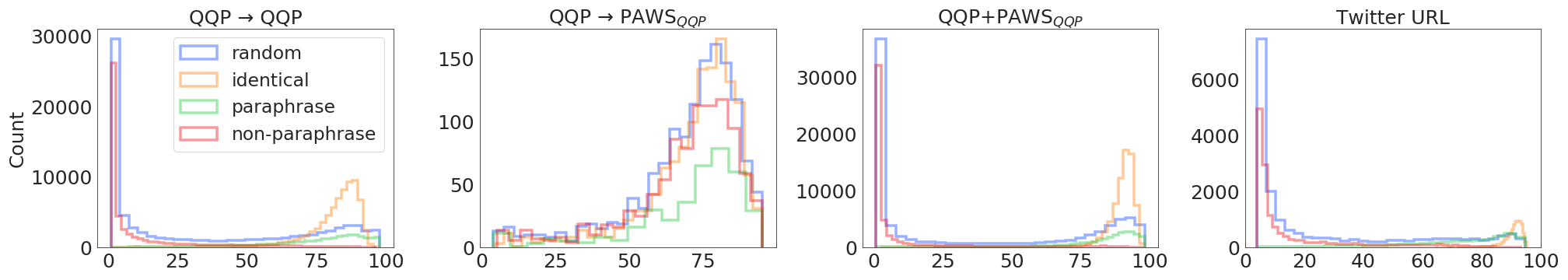}
    \caption{}
    \label{confidence_score_hist}
\end{subfigure}

    \caption{Histograms of (a) the score difference between randomly-selected and identical pairs and (b) paraphrase score for sentence pairs. Randomly-selected pairs contain the sentence pairs their original and reverse order. (We do not include the plot for MRPC since most paraphrase pairs from the dataset look alike, and it is hard to distinguish the distributions from the graph.)}
    \label{confidence_kde}
\end{figure*}

\subsection{Random Sentences}
\label{random-vs-identical}
We augmented the original datasets with sentence pairs concatenated in opposite order, as in Section~\ref{sentence_order}, and labeled them same as their original pairs. We then compared each sentence pair, $(s,s^{\prime})$, to a pair of identical sentences, $(s,s)$, given the same query sentence $s$. We fine-tuned BERT on each dataset to learn a paraphrase score function $f$, and computed the fraction of tests where a randomly-selected pair gets a higher paraphrase score than an identical pair, $f(s,s^{\prime}) > f(s,s)$. Table~\ref{paraphrase_score} shows the results, revealing a similar pattern as in Section~\ref{identifying_identical_pairs}. The model trained on QQP considers more than 30\% of randomly-selected paraphrase sentence pairs to be more similar than the identical pairs, but the ratio decreases to 21\% when adding the adversarial set into training. For MRPC and Twitter URL corpus, less than 5\% of paraphrase pairs are considered to be more similar than the identical pairs.

For a randomly-selected sentence pair, $(s, s^{\prime})$, and a pair of identical sentences, $(s, s)$, given the same query $s$ sentence, we computed the score difference as $f(s, s^{\prime}) - f(s, s)$. The distributions of the score differences are shown in Figure~\ref{confidence_diff_hist}. We filtered out the pairs that have lower paraphrase score than the identical pairs, and report the average score difference in Table~\ref{paraphrase_score}. In Figure~\ref{confidence_diff_hist}, the model trained on QQP has the largest score difference between paraphrase and identical pairs. After augmenting the training set with PAWS\textsubscript{QQP}, the right tail of the distribution for paraphrase pairs diminishes. This indicates that the model considers fewer non-identical sentences as more similar to the query sentence than itself.

\subsection{Paraphrase Score Distribution}
To better understand how the scores are distributed, we plot the histograms of paraphrase score for random, paraphrase, non-paraphrase, and identical sentence pairs in Figure~\ref{confidence_score_hist}. In the normal setting, there are two peaks in the distributions of randomly-selected pairs since they include both paraphrase and non-paraphrase pairs. On the other hand, the sentence pairs from PAWS\textsubscript{QQP} all seem very similar to the model. The distributions clearly show the model cannot distinguish them. Compared with the distribution for the Twitter corpus, the distribution of paraphrase pairs from QQP is more spread out, and it has slightly larger gap between the distribution of paraphrase and identical pairs.

\section{Discussion}

\noindent\textbf{Defining Paraphrases.} Our experiments assume that the ``best'' paraphrase for a given sentence $s$ is $s$ itself. This assumes an equivalent in meaning definition of paraphrase, but other definitions may be appropriate. \citet{what-is-a-paraphrase} defined paraphrases as ``sentences that convey the same meaning using different wording". By this definition, identical sentences are not paraphrases. 
Of course, we do not need a complex model to identify identical sentences when a simple equality test will do. However, when considering paraphrase detection as a test for how well language models can understand meaning, it would be counterproductive to consider identical sentences non-paraphrases, and require a trivial modification to consider them perfect paraphrases. Thus, we would expect a model to be able to identify sentence pairs with the same meaning as paraphrases regardless of whether they are the same in their surface forms.

Our experiments also assume that the paraphrase relationship should be symmetrical. This is consistent with the notion that the paraphrase identification task is meant to identify sentences with similar meaning, but not consistent with the purpose of many uses of paraphrase identification (e.g., in some real world question retrieval tasks, finding questions that contain the query, or that have the opposite meaning, would still be useful). 
This suggests the importance of a clear notion of what a paraphrase is, in both constructing test datasets and in determining how a given application can use a paraphrase detection model.


\noindent\textbf{Selection Bias in the Pointwise Setting.}
Previous studies have addressed the problem of selection bias when constructing the task as a pointwise learning problem \citep{l2r-selection-bias,zadronzny-bianca-selection-bias}. Datasets tend to have inconsistent frequency of sentences causing the model biased towards the dominating sentences. For instance, we found that some sentences from the Twitter corpus are repeated almost a hundred times as the first input sentence. This is part of the reason that the model gets more asymmetrical prediction results for sentences in reverse order (Table~\ref{asymmetry_ratio}). 

\section{Conclusion}
Although the state-of-art paraphrase identification models can achieve impressive performance under the pointwise evaluation method, they cannot handle real-world problems and unseen data well and even have worse results than a BOW model on simple tasks. We show that the asymmetry in BERT can produce inconsistent prediction results when reversing the order of the two sentences. We examined the relation of semantic equivalence learned by models trained with pointwise approach, and found that they may consider a random sentence as more similar to the query sentence itself. This suggests future work to reconsider how to match the training and evaluation to the actual objective of downstream applications, and thus create more reliable evaluation metrics and benchmarks.

\bibliography{acl2020}
\bibliographystyle{acl_natbib}

\end{document}